\documentclass[conference]{IEEEtran}
\IEEEoverridecommandlockouts
\usepackage{cite}
\usepackage{amsmath,amssymb,amsfonts}
\usepackage{float}
\usepackage{algpseudocode}
\usepackage[ruled,linesnumbered]{algorithm2e}
\usepackage{graphicx}
\usepackage{textcomp}
\usepackage{xcolor}
\usepackage{booktabs}  %三线表
\usepackage{caption}
\usepackage[]{subfig}
\usepackage[colorlinks]{hyperref} % 颜色可调
\usepackage[numbers,sort&compress]{natbib}
\usepackage{amsthm}
\theoremstyle{remark}

\theoremstyle{definition}
\newtheorem{theorem}{Theorem}

\def\BibTeX{{\rm B\kern-.05em{\sc i\kern-.025em b}\kern-.08em
    T\kern-.1667em\lower.7ex\hbox{E}\kern-.125emX}}
\begin{document}
\captionsetup{font={footnotesize}}

\title{Clustered Data Sharing for Non-IID Federated Learning over Wireless Networks}%\\{\footnotesize \textsuperscript{*}Note: Sub-titles are not captured in Xplore and should not be used}\thanks{Identify applicable funding agency here. If none, delete this.}}
\author{\IEEEauthorblockN{Gang Hu\textsuperscript{1}, Yinglei Teng\textsuperscript{1} \textit{Senior Member, IEEE}, Nan Wang\textsuperscript{1}, F. Richard Yu\textsuperscript{2}, Fellow, IEEE}
	\textsuperscript{1}Beijing Key Laboratory of Space-ground Interconnection and Convergence\\
Beijing University of Posts and Telecommunications (BUPT), Xitucheng Road No.10, Beijing, China, 100876.\\
    \textsuperscript{2}Department of Systems and Computer Engineering, Carleton University, Ottawa, ON K1S 5B6, Canada\\

	Email: hugang@bupt.edu.cn, lilytengtt@bupt.edu.cn, wangnan\_26@bupt.edu.cn, richard.yu@carleton.ca \\
}

%\author{\IEEEauthorblockN{1\textsuperscript{st} Yangliu Zhao}
%\IEEEauthorblockA{\textit{School of Electronic Engineering} \\
%\textit{Beijing University of Posts and Telecommunications}\\
%Beijing, China \\
%Email: zhaoyangliu@bupt.
%edu.cn}
%\and
%\IEEEauthorblockN{2\textsuperscript{nd} Yinglei Teng}
%\IEEEauthorblockA{\textit{School of Electronic Engineering} \\
%\textit{Beijing University of Posts and Telecommunications}\\
%Beijing, China \\
%Email: lilytengtt@bupt.edu.cn}
%\and
%\IEEEauthorblockN{3\textsuperscript{rd} An Liu}
%\IEEEauthorblockA{\textit{College of Information Science and Electronic Engineering} \\
%\textit{Zhejiang University}\\
%Hangzhou, China \\
%Email: anliu@zju.edu.cn}
%\and
%\IEEEauthorblockN{4\textsuperscript{th} Vincent Lau}
%\IEEEauthorblockA{\textit{Department of Electronic and Computer Engineering} \\
%\textit{Hong Kong University of Science and Technology}\\
%Hong Kong, China \\
%Email: eeknlau@ece.ust.hk}
%%\and
%%\IEEEauthorblockN{4\textsuperscript{th} Vincent Lau}
%%\IEEEauthorblockA{\textit{dept. name of organization (of Aff.)} \\
%	%\textit{name of organization (of Aff.)}\\
%%	City, Country \\
%%	email address or ORCID}
%}
\maketitle
\begin{abstract}
Federated Learning (FL) is a novel distributed machine learning approach to leverage data from Internet of Things (IoT) devices while maintaining data privacy. However, the current FL algorithms face the challenges of non-independent and identically distributed (non-IID) data, which causes high communication costs and model accuracy declines. To address the statistical imbalances in FL, we propose a clustered data sharing framework which spares the partial data from cluster heads to credible associates through device-to-device (D2D) communication. Moreover, aiming at diluting the data skew on nodes, we formulate the joint clustering and data sharing problem based on the privacy-preserving constrained graph. To tackle the serious coupling of decisions on the graph, we devise a distribution-based adaptive clustering algorithm (DACA) basing on three deductive cluster-forming conditions, which ensures the maximum yield of data sharing. The experiments show that the proposed framework facilitates FL on non-IID datasets with better convergence and model accuracy under a limited communication environment.
\end{abstract}


\section{Introduction}
\makeatletter
\newcommand{\rmnum}[1]{\romannumeral #1}
\newcommand{\Rmnum}[1]{\expandafter\@slowromancap\romannumeral #1@}
\makeatother
Due to the proliferation of advanced sensors in wireless networks, a massive volume of data (e.g. images, videos, and voices) is continuously generated in Internet of Things (IoTs) \cite{1}. This big data is usually offloaded and stored in the cloud, together with Deep Learning (DL) approaches to enable multifarious intelligent services. However, the privacy leakage problem becomes critical when the raw sensitive data is exposed to attacks and cyber risks, making data owners reluctant to share their data. Hence, a novel distributed learning paradigm called Federated Learning (FL) is proposed to train a high quality model using locally-trained models instead of local data \cite{2}.

%All devices in the FL system collectively reap the benefits of shared models trained from the rich data, without the need to centrally store it \cite{3}. In particular, all devices iteratively train a local DL model using their private local data and forward their models to a central server. The server then aggregates them into a global model, and sends it back to devices for the subsequent round of training. In this way, the FL architecture allows devices to only share their trained models thus avoid revealing their data.
In the framework of FL, a number of federated rounds are required between clients and server to coordinately achieve a target accuracy \cite{1}. Due to the millions of parameters in the complex deep learning model, updating the high-dimensional model would result in significant communication costs. However, this case is more serious when data across the devices is non-independent and identically distributed (non-IID). Practically, the data on distributed devices are differentially generated due to confined observation ability, leading to the imbalance of local data samples. According to research studies \cite{4,16}, the higher the degree of non-IID in the data, the slower convergence rate of the FL algorithm. Therefore, the heterogeneous data based on learning coordination becomes the bottleneck of FL when applied in the actual IoT environment.

There is plenty of literature attempting to address the statistical heterogeneity challenge. One approach is to design effective FL algorithms under non-IID data \cite{6,15,7}. Li \emph{et al}. \cite{6} add a proximal term in each local objective function to limit the distance between the local model and the global model. FedNova \cite{15} provides a normalized averaging method that eliminates objective in-consistency while preserving fast error convergence. SCAFFOLD \cite{7} controls the variance among the participants to correct the update direction in local training.
Another approach aims at augmenting the local data by making data across clients closer to the IID case. In \cite{9}, the server shares a small amount of data with all devices so that the distance between data distributions decreases. \cite{10} designs a joint optimization algorithm to balance the model accuracy and the cost based on data sharing method on wireless networks. There are other works gathering the local data information on the server by collecting the uploaded data \cite{5} or distilled knowledge \cite{8}.
However, the former algorithm improving techniques obtains very limited elevation of the model accuracy on non-IID data, while the latter data augment based methods are usually expensive and insecure since they need to create a publicly available proxy data source on the server side.

Unlike the previous works, we propose a framework which reduces the degree of non-IID data and does not rely on a central server or even additional datasets. This framework solves the non-IID challenge by selecting some devices as cluster heads to share partial data with trustable associates. In non-IID FL, devices with high-quality data can accelerate FL training. Employing them as cluster heads, the data heterogeneity can be largely shrunk by data sharing. Moreover, the communication overhead and privacy issues can be well regulated by sparing data with trustworthy and reliable partners in the Device-to-Device (D2D) multicast technology.

The contributions of this paper are summarized as follows: A clustered data sharing framework is proposed to evaluate and eliminate the impacts of distribution skew on the convergence of FL and target accuracy. Furthermore, this framework can be applied to most non-IID FL algorithms as a data preparation procedure. We formulate the problem of minimizing the distribution distance under privacy and communication constraints. We theoretically analyze this complicated optimization problem from individual, intra-cluster, inter-cluster perspectives and design a clustering method called distribution-based adaptive clustering algorithm (DACA) that adaptively selects users based on the constraint graph. Experimental results show that the scheme effectively improves the accuracy of non-IID FL and reduces the number of communication rounds.
\begin{figure}[t]
\centering
\includegraphics[width=8.2cm,height=5.1cm]{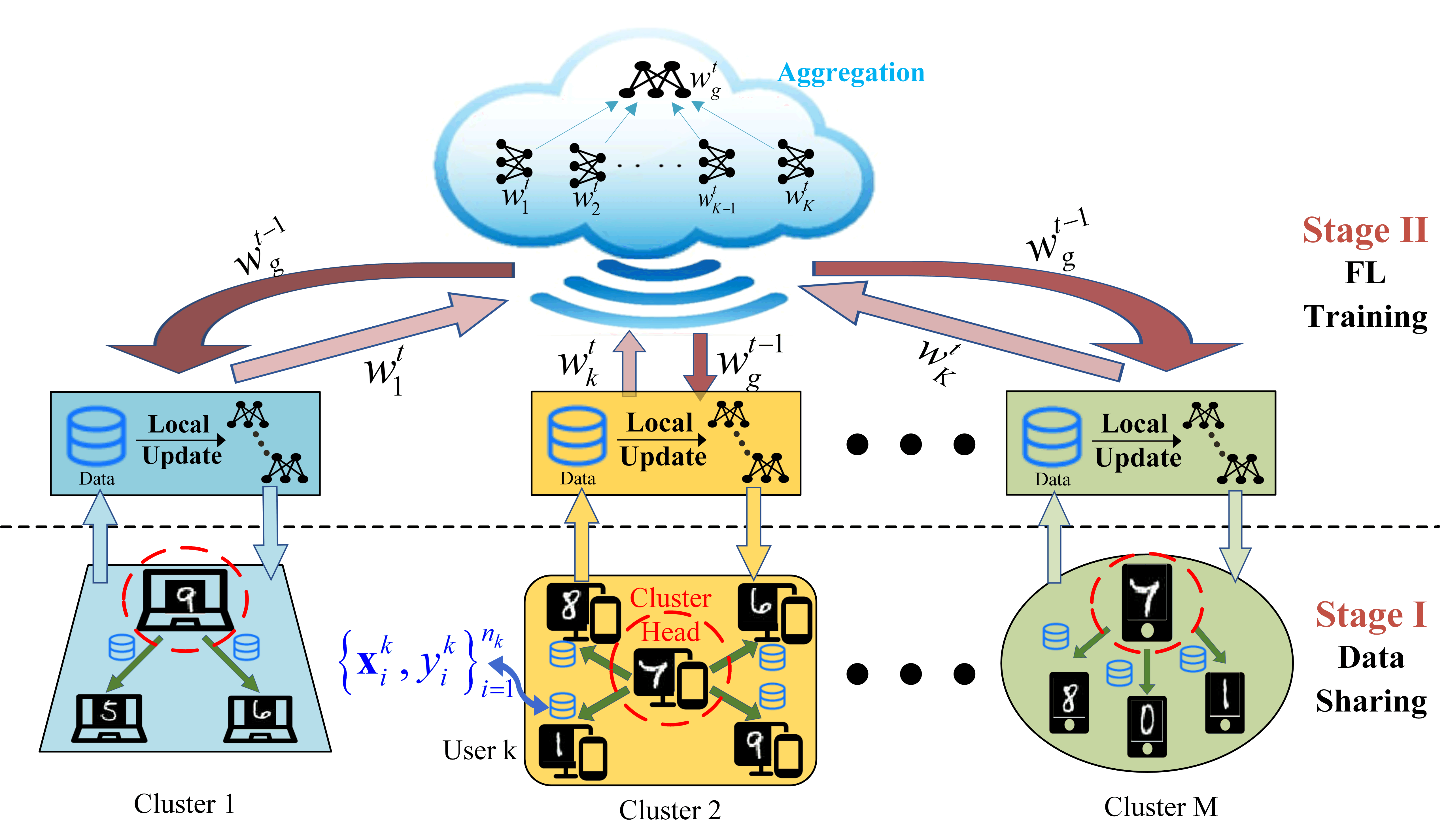}
\caption{Clustered data sharing framework for FL.}
\vspace{-5mm}
\label{fig_1}
\end{figure}
\section{System Model}\label{SysMod}
Consider a wireless edge computing system consisting of a base station (BS) and a set of $K$ users (i.e., smart mobile devices) denoted as ${\cal K} = \{ 1,2,...,K\} $ as in Fig. \ref{fig_1}. Each user collects the local samples ${{{\cal N}}_k} = \{ {\bf{x}}_i^k,y_i^k\} _{i = 1}^{{n_k}},\;k \in {{\cal K}}$, where ${\bf{x}}_i^k$ is the input vector of sample $i$, $y_i^k \in \left\{ {1,...,Y} \right\}$ is the corresponding labels, and ${n_k}$ is the number of the sample-label pairs. For the DL task, e.g., image classification, the users and the BS work collaboratively under client-server architecture. In this edge environment, besides the common downlink/uplink transmission, D2D communications are allowed for users in close proximity.
\subsection{Non-IID Data Model for FL}\label{NonIIDMod}
The local data generated at clients are statistically different, and may vary significantly, resulting in the typical non-IID cases. For clarity, the mathematical definition of IID and non-IID is provided as below. IID is a generally known assumption in distributed learning, i.e., $\left\{ {{\bf{x}}_i^k,y_i^k} \right\}_{i = 1}^{{n_k}}$ follows the global distribution ${P_g}({\bf{x}},y)$. Nevertheless, the FL violates the IID assumption, since typically the users only have local data due to the limitations of the geo-regions and observation ability. For non-IID setting, the dataset of a particular user $\left\{ {{\bf{x}}_i^k,y_i^k} \right\}_{i = 1}^{{n_k}}$ follows a distinct distribution ${P_k}\left( {{\bf{x}},y} \right)$ which can be rewritten as ${P_k}\left( y \right){P_k}\left( {{\bf{x}}\left| y \right.} \right)$. In this case, even though ${P_k}({\bf{x}}\left| y \right.)$ is the same across users, the marginal distribution ${P_k}(y)$ could differ. This is the common non-IID case called \emph{Label distribution skew} \cite{6,15,7,9}. There are some other non-IID cases related to feature distributions ${P_k}({\bf{x}})$, conditional distribution ${P_k}({\bf{x}}\left| y \right.)$ or ${P_k}(y\left| {\bf{x}} \right.)$. Like the majority of non-IID FL works, we only focus on \emph{Label distribution skew} ${P_k}(y)$ in this paper.

Intuitively, the impact of non-IID data can be portrayed by the deviation between the model ${w_{\rm{g}}}$ trained on non-IID data and ${w_{{\rm{IID}}}}$ trained on IID data. However, the weight information will only be known after FL is completed. Since model training is data dependent, the model deviation can be quantified in terms of the distance between ${P_k}(y)$ and ${P_g}(y)$. There are several commonly used methods to measure the distance between two distributions, e.g., ${l_p}$-norm, Kullback-Leibler (KL) divergence. However, there is no theoretical connection between these methods and model deviation. We derive a term by analyzing the model deviation $\left\| {{w_g} - {w_{{\rm{IID}}}}} \right\|$\footnote{The detailed process of this inference can be found in \cite{9}},
\vspace{-3mm}
\begin{equation}\label{Eq:1}
	{\left\| {w_g \hspace{-1mm} - \hspace{-1mm} w_{{\rm{IID}}}} \right\| \hspace{-1mm} \propto \hspace{-1mm} \sum\limits_{i = 1}^Y {\left\| {{P_k}\left( {y = i} \right)\hspace{-1mm} - \hspace{-1mm} {P_g}\left( {y = i} \right)} \right\|} \hspace{-1mm} \buildrel \Delta \over = \hspace{-1mm}{D_{{\rm{EMD}}}}\left( k \right)},
\end{equation}
where ${D_{{\rm{EMD}}}}\left( k \right)$ is the earth mover’s distance (EMD) which can be used to quantify the degree of non-IID data on device $k$. Moreover, we define the system EMD as the weighted sum of separate EMD values
\begin{equation}\label{Eq:2}
	{{\bar D_{{\rm{EMD}}}} = \sum\limits_{k = 1}^K {\frac{{{n_k}}}{n}\sum\limits_{i = 1}^Y {\left\| {{P_k}\left( {y = i} \right) - {P_g}\left( {y = i} \right)} \right\|} }},
\end{equation}
where $n = \sum\nolimits_{k = 1}^K {{n_k}} $ is the size of all distributed samples.

\begin{figure}[!t]
\centering
\includegraphics[width=6cm,height=3.5cm]{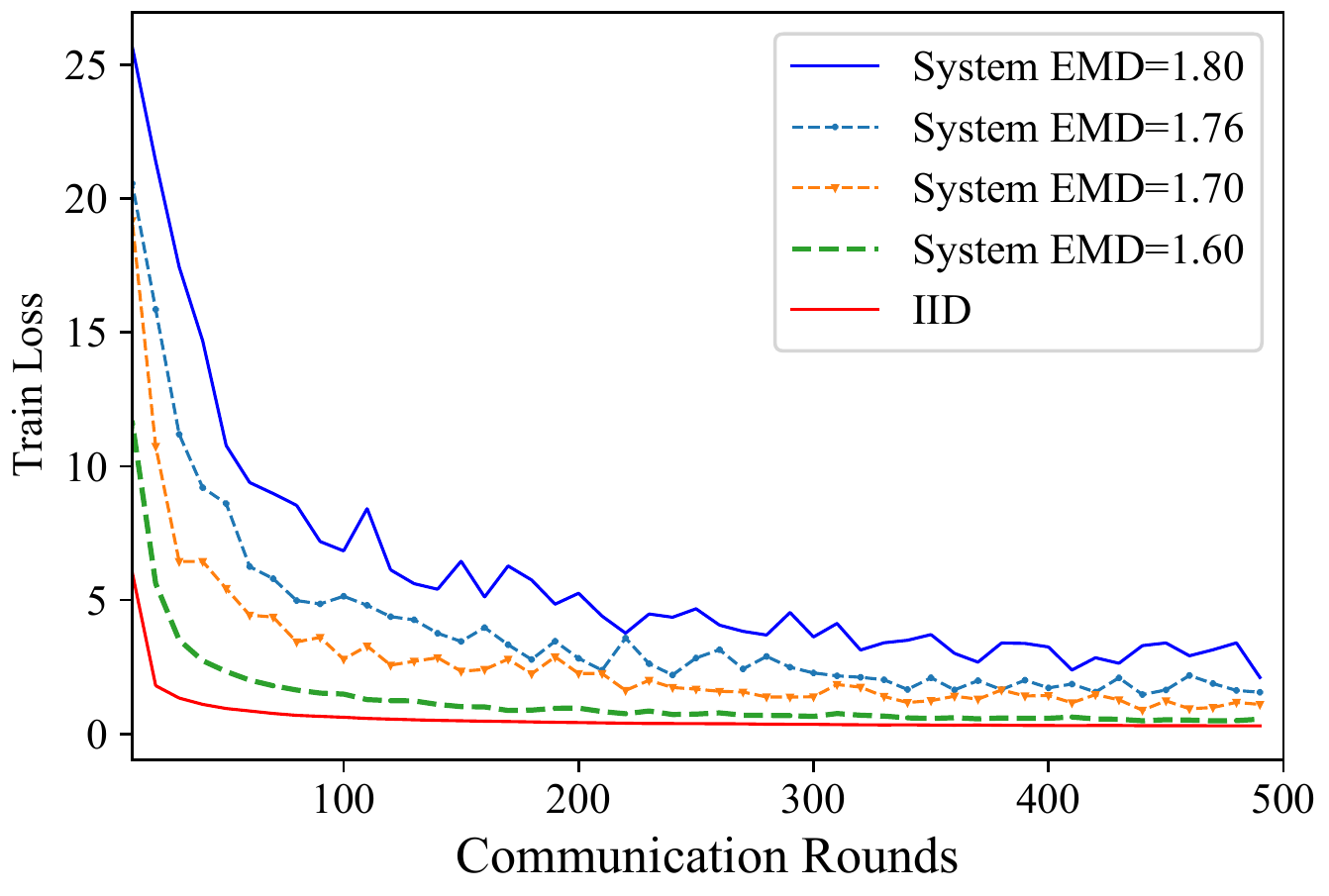}
\caption{Training loss over communication rounds of FL compared to different non-IID data. There are 100 users training CNN models on the MNIST dataset, and we use the same model architectures and hyperparameter setting as \cite{3}.}
\vspace{-5.7mm}
\label{fig_2}
\end{figure}

To illustrate the effect of the system EMD, we conduct an experiment to train FL model on diverse non-IID cases, and the results are shown in Fig. \ref{fig_2}. It is observed that the smaller the system EMD, the lower training loss and faster convergence rate it brings, which indicates that mitigating the system EMD can effectively accelerate FL training and improve model accuracy.

\subsection{Clustered Data Sharing FL Framework}\label{FLMod}
 To mitigate the data unbalance feature for FL, we propose a communication-aware clustered data sharing scheme which makes data distribution across devices more homogeneous by exchanging a small amount of data within a communication-efficient and privacy-preserving cluster. As shown in Fig. \ref{fig_1}, this framework consists of two stages, i.e.,

\textbf{Data preparing stage:} during this stage, users are clustered with respect to data distribution and other factors (i.e., channel states, credibility), where the cluster heads share a subset of their own data to cluster members through reliable multicast D2D communication. Accordingly, each node can compute their local parameters with the mixed samples. Denote the cluster head $m \in {\cal M}$ and cluster members ${{\cal C}_m}$. Note that one node cannot join in more than one cluster so as to avoid conflicts, i.e., ${{\cal C}_m} \cap {{\cal C}_l} = \emptyset ,\forall m,l \in {\mathcal M},m \ne l$. With a copy of shared data, the local data volume on the device $k$ becomes
\begin{equation}\label{Eq:3}
	{{\tilde n_k} = \left\{ \begin{array}{l}
{n_k} + n_m^s, \; \; k \in {{\cal C}_m}\\
{n_k},\quad\quad\quad\,{\rm{      }}k \notin {{\cal C}_m}{\rm{    }}
\end{array} \right.},
\end{equation}
where $n_m^s$ is the data volume of sharing from cluster head $m$. Without loss of generality, we specify ${P_m}\left( y \right) = {P_k}\left( y \right)$ if $k \notin {{\cal C}_m}$ and the new distribution of the user $k$ can be derived as
\begin{equation}\label{Eq:4}
	{{\tilde P_k}\left( y \right) = \frac{{{n_k}{P_k}\left( y \right) + n_m^s{P_m}\left( y \right)}}{{{n_k} + n_m^s}}}.
\end{equation}
Intuitively, data sharing can modify the original data distribution on devices, diluting the disparity in data feature. Therefore, the value of the system EMD after data sharing is given by
\vspace{-4mm}
\begin{equation}\label{Eq:5}
	{{\tilde D_{{\rm{EMD}}}} = \sum\limits_{k = 1}^K {\frac{{{{\tilde n}_k}}}{n}\sum\limits_{i = 1}^Y {\left\| {{{\tilde p}_k}\left( {y = i} \right) - {p_g}\left( {y = i} \right)} \right\|} } }.
\end{equation}

\textbf{The FL training stage:}  is to train a global model via collaborations between the server and multiple clients, where the global model is obtained by minimizing the following loss function:
\vspace{-4mm}
\begin{equation}\label{Eq:6}
    {\mathop {\min }\limits_{{w_g}} \frac{1}{K}\sum\limits_{k = 1}^K {\frac{1}{{{n_k}}}\sum\limits_{i = 1}^{{n_k}} {{F_k}\left( {{\bf{x}}_i^k,y_i^k,{w_g}} \right)} } }.
\end{equation}
Here, ${F_k}\left(  \cdot  \right)$ is the loss function for user $k$. In particular, the model is updated with the following two steps during the $t$-th round:

\emph{1) Local model training:} With the global model parameters $w_g^{t - 1}$ from the BS, users train local models $w_k^t$ separately using their own local dataset $\left\{ {{\bf{x}}_i^k,y_i^k} \right\}_{{\rm{i = 1}}}^{{n_k}}$ by stochastic gradient descent (SGD)
\vspace{-3mm}
\begin{equation}\label{Eq:7}
    {w_k^t = w_g^{t - 1} - \eta \nabla {F_k}\left( {{x_i},{y_i};w_g^{t - 1}} \right)},
\end{equation}
where $\eta $ is the learning rate in SGD, and $\nabla {F_k}\left(  \cdot  \right)$ is the gradient of loss function ${F_k}\left(  \cdot  \right)$ with respect to $w_g^{t - 1}$.

\emph{2) Global model aggregation:} When the local training step is accomplished, the users send the local models to the BS for synchronous aggregation. By averaging the uploading models from all users, the global model $w_g^t$ is updated as
\vspace{-3mm}
\begin{equation}\label{Eq:8}
    {w_g^t = \sum\limits_{k = 1}^K {\frac{{{n_k}}}{n}w_k^t} }.
\end{equation}

After global model aggregation, $w_g^t$ is fed back to the users, and the above FL training procedure repeats for multiple rounds until convergence.

\subsection{Data Sharing Model}
As is known, data sharing is unavoidably accompanied by privacy risk and communication costs. To mitigate these effects, two metrics are defined for further design of cluster algorithm.

\textbf{Social closeness:} is one of the factors of social awareness referring to users having different levels of trust with each other. Regarding data privacy, users' reluctance to expose data is weakened for close social relationships. For clustered data sharing, we consider the social closeness ${e_{m,c}}$ between the cluster head $m \in M$ and cluster member $c \in {{\cal C}_m}$. To ensure trustable data exchange, the cluster members are required to maintain a socially closed affiliation with the cluster head, i.e., ${e_{m,c}} \ge {e_{{\rm{th}}}},\forall m,c$. For clarity, we build a graph ${{\cal G}} = \left( {{{\cal K}},{{\cal E}}} \right)$ to more illustrate the social closeness among users, where ${{\cal E}} = \{ {e_{k,j}} \in \left[ {0,1} \right],\forall k,j \in {{\cal K}}\} $ is the set of graph edges.

\textbf{Transmission delay:} The cluster head shares part of datasets with the group members via D2D multicast communications. The transmission rate of data sharing from the cluster head $m$ to each cluster member $c \in {{{\cal C}}_m}$ is given by
\vspace{-2mm}
\begin{equation}\label{Eq:9}
	{{v_{m,c}} = B_m^s{\log _2}\left( {1 + \frac{{P_m^s{h_{m,c}}}}{{{I_m} + B_m^s{N_0}}}} \right)},
\vspace{-0.5mm}
\end{equation}
where $B_{m}^s$ is the bandwidth that the cluster head $m$ used to multi-broadcast the partial dataset; and ${h_{m,c}}$ is the channel gain between cluster head $m$ and member $c$. $P_m^s$ is the transmit power of cluster head, and ${N_0}$ is the noise power spectral density. ${I_m}$ is the inference caused by other clusters which are located in other service areas. With the multicast, the delay for data sharing within one cluster is determined by the worst link. Meanwhile, the total data sharing time cost ${\tau ^s}$ depends on the maximal delay of clusters, and is thereby given by
\vspace{-3mm}
\begin{equation}\label{Eq:10}
	{\tau _{}^s = \mathop {\max }\limits_{m \in {{\cal M}},c \in {{\cal C}_m}} \left. {\left\{ {\frac{{an_m^s}}{{{v_{m,c}}}}} \right.} \right\}},
\end{equation}
where $a$ denotes bits of each data sample.
\section{Problem Formulation And Solution}\label{SOLUTIONS}

\subsection{Problem Formulation}
In clustered data sharing FL framework, an appropriate clustering method involving cluster head selection should guarantee the training of the high-quality global model in a communication-efficient way. As shown in Sec. \ref{SysMod}, the main culprit damaging non-IID FL training can be quantified by EMD. Therefore, we design the optimization framework that minimizes the system EMD with joint design of data sharing and clustering strategy while guaranteeing  guarantee users' privacy and low communication costs as follows:
\vspace{-3mm}
\begin{align}
{\mathcal{P}:} \quad \mathop { \min }\limits_{{\cal M},{{\cal C}},{{\bf{N}}^s}} {\tilde D_{{\rm{EMD}}}},\quad\quad\quad\quad\quad\quad\quad\quad\quad \label{Eq:11}\\
{s.t. \quad {{{\cal C}}_m} \cap {{{\cal C}}_l} = \emptyset ,\;\;\forall m,\;l \in {{\cal M}},\;m \ne l,} \tag{11a} \label{11a}\\
{n_m^s \le {n_m},\quad \forall m \in {{\cal M}},\quad\quad\quad\quad\quad} \tag{11b} \label{11b}\\
{{e_{m,c}} \ge {e_{th}},\;\;\;\forall m \in {{\cal M}},\;c \in {{{\cal C}}_m ,\quad}\tag{11c}} \label{11c} \\
{{\tau ^s} \le {T_{th} ,\;\quad\quad\quad\quad\quad\quad\quad\quad\quad\quad\;}} \tag{11d} \label{11d}
\end{align}
where ${{\cal C}}{\rm{ = }}\left\{ {\left. {{{{\cal C}}_1} \ldots ,{{{\cal C}}_m}, \ldots } \right\}} \right.$ and ${{{\cal C}}_m}$ is the set of cluster members connected with cluster head $m$, ${{\bf{N}}^s} = \left[ {n_1^s, \ldots ,n_m^s, \ldots } \right]$ is a vector of data volume for sharing from the cluster heads. (\ref{11a}) is the constraint for exclusive cluster forming. (\ref{11b}) confines the maximum data sharing volume. The constraint (\ref{11c}) restricts the privacy requirement for data sharing between cluster heads and cluster members.  In our framework, to maintain a proper sharing preparing duration, we set the  maximal limit for transmission delay as (\ref{11d}).

\subsection{Problem Analysis}

The data sharing strategy to tune the non-IID degree is powerful yet intractable due to the following reasons. \emph{Firstly}, the joint decision of clustering and sharing data size is coupled and the relationship between variables and objective is implicit due to no-closed expression. \emph{Secondly}, the clustering and cluster head selection constructs an NP-hard problem that existing practices all rely on heuristic algorithms.  \emph{Moreover}, the cluster forming incorporates restrictions of privacy preserving and transmission efficiency as well, making the problem further intricate. To identify the complex clustering effect on $\tilde D_{\rm{EMD}}$, we first put the external constraints (\ref{11b})-(\ref{11d}) asides. Next, we devise the following three conditions to analyze the reward of data sharing on $\tilde D_{\rm{EMD}}$ from the perspectives of individual, intra-cluster and inter-cluster.

\textbf{Condition 1: (individual perspective)}
After data sharing, the EMD of an arbitrary user should be as small as possible, i.e.,
\vspace{-4mm}
\begin{equation}\label{Eq:13}
	{\mathop {\min }\limits_{\forall k \in {{\cal K}}} \sum\limits_{i = 1}^Y {\left\| {{{\tilde p}_k}\left( {y = i} \right) - {p_g}\left( {y = i} \right)} \right\|} }.
\end{equation}
Especially, if {\small $\sum\nolimits_{i = 1}^Y {\left\| {{{\tilde p}_k}\left( {y = i} \right) - {p_g}\left( {y = i} \right)} \right\|} \hspace{-1mm} =\hspace{-1mm} 0$}, the data turns into the ideal IID. Without considering any constraint, condition 1 is equivalent to the original optimization problem. But it is still hard to meet this condition, and thus we convert the target into the following two extended conditions:

%\addtolength{\topmargin}{0.04in}
\textbf{Condition 2: (intra-cluster perspective)}
Within a cluster, a best data sharing is secured if the EMD of the cluster head and cluster members differs as much as possible, i.e.,
\vspace{-3mm}
\begin{equation}\label{Eq:14}
	{\mathop {\max }\limits_{\forall c \in {{\cal C}_m}} {D_{{\rm{EMD}}}}\left( c \right) - {D_{{\rm{EMD}}}}\left( m \right)}.
\end{equation}
Regarding the EMD of cluster members is already given, to release the divergence after data sharing, the EMD of cluster heads should be as low as possible. To put it briefly, nodes with high data quality is more suitable for the cluster heads. Although condition 2 ensures the best benefit for sharing within a cluster, the clustering margin is very unclear, e.g., how many clusters would be the best. Thus, additional requirements should be enforced to define the inter-cluster relationship.

\textbf{Condition 3: (inter-cluster perspective)}
Given the results of a particular clustering ${\cal M}$ and ${\cal C}$, there exists a better clustering method if the distribution distance of any member $c \in {{\cal C}_m}$ and other cluster head $m'$ is larger than that with the current head $m$, i.e.,
\vspace{-3mm}
\begin{equation}\label{Eq:15}
%{\mathop {\max }\limits_{m \in {\cal M}} {D'_{{\rm{EMD}}}}\left( {k,m} \right)}.
{{D_{{\rm{EMD}}}}\left( {c,m} \right) \ge {D_{{\rm{EMD}}}}\left( {c,m'} \right),\forall c \in {{\cal C}_m},m \ne m'}.
\end{equation}
Here,{\small{ ${D_{{\rm{EMD}}}}\left( {c,m} \right)\hspace{-1mm}=\hspace{-1mm}\sum\nolimits_{i = 1}^Y {\left\| {{P_c}\left( {y = i} \right)\hspace{-1mm} - \hspace{-1mm} {P_m}\left( {y = i} \right)} \right\|}$}} is the EMD between data distributions on cluster member $c$ and the cluster head $m$. For the critical nodes that can belong to multiple clusters, this condition can be used as a separation criterion. An intuitive interpretation is that the nodes will join in the cluster whose data distribution of cluster head differ much from their own.

%\begin{remark}
%  \cite{14} groups the devices into multiple clusters that perform FL cyclically in each round and the convergence rate depends on the data heterogeneity of %intra-cluster data.
%\end{remark}

%\begin{remark}
%  \cite{12,17} trains the multi-models in clustering users by aggregating the local models from the same cluster. In this case, the data heterogeneity is from the inter-cluster data.
%\end{remark}

Aided by the condition 2 and condition 3, the solutions to the relaxed problem $\mathcal{P}$ can be obtained by the following theorem.

\begin{theorem} \label{thm}

\emph{Putting the constraints (\ref{11b})-(\ref{11d}) aside, if ${{\cal M}^*}$ and ${{\cal C}^*}$ are the optimal solutions to  the problem $\mathcal{P}$, then condition 2 and condition 3 both hold with ${{\cal M}^*}$ and ${{\cal C}^*}$.}

%\emph{If condition 2 and condition 3 both hold with ${{\cal M}^*}$ and ${{\cal C}^*}$, then ${{\cal M}^*}$ and ${{\cal C}^*}$ are the optimal solutions of problem %(\ref{Eq:11}).}

\end{theorem}
\vspace{-3mm}
\emph{proof.} See Appendix \ref{proof_1}.
\vspace{-1mm}

\subsection{Distribution Based Adaptive Clustering Algorithm}
Incorporating the constraints (\ref{11b})-(\ref{11d}), the sharing data volume within a specified time period and the privacy threshold among associates both affect the problem $\mathcal{P}$, making Theorem \ref{thm} not immediately applicable. The constraint (\ref{11b}) plays the role of the upper bound on sharing data volume, which has almost no relevance to the clustering result. The constraint (\ref{11c}) associates with the connectability with credible nodes. The delay for data sharing in (\ref{11d}) directly relates to the sharing data and  the transmission rate according to Eq. (\ref{Eq:10}), thus constraint (\ref{11d}) can be transformed to additional restriction to sharable data size and an equivalent rate threshold constraint ${v_{{\rm{th}}}}$. Following the above analysis, we reform a constrained graph {\small ${\cal G}' \hspace{-1mm} = \hspace{-1mm} \left( {{\cal K},{\cal E}'} \right)$}, where {\small ${\cal E}' = \left\{ {{{\tilde e}_{k,j}}= {D_{{\rm{EMD}}}}\left( {k,j} \right)|{e_{k,j}} \ge {e_{{\rm{th}}}},{v_{k,j}} \ge {v_{{\rm{th}}}}} \right\}$}.

As shown in Algorithm \ref{algorithm}, we propose a distribution-based adaptive clustering algorithm (DACA) by applying the Theorem \ref{thm} to the privacy-preserving and communication-efficient constrained  graph {\small ${\cal G}' \hspace{-1mm} = \hspace{-1mm} \left( {{\cal K},{\cal E}'} \right)$}. According condition 2 and condition 3, the solution of the problem $\mathcal{P}$ can be transformed into portable operations by selecting nodes with low EMD as few as possible to cover the whole graph. This transformation can be easily explained: if the clusters are not the least, it means that a nodes with a higher EMD is selected as the cluster head, which contradicts the condition 3. The specific procedure includes two steps, cluster heads selection and cluster members association. For the cluster heads selection, calculate the ${D_{{\rm{EMD}}}}\left( k \right)$ and sort it, then the cluster heads can be selected in descending order until all nodes are covered. For the cluster members, each node is allowed to associate with the best head within the constrained graph ${\cal G}'$. Meanwhile, the data sharing volume ${{\bf{N}}^s}$ is calculated from constraints (\ref{11b}) and (\ref{11d}) after the cluster is determined. With this algorithm, the complicated cluster form is optimally approached with very low complexity, and the majority of the computational consumption lies in the generation of the constraint graph. Moreover, the proposed DACA algorithm groups nodes adaptively without relying on the clustering number as an priori parameter. It is worth noting that this algorithm can be well applied to similar constraints confined clustering problems by resolving the original objective.

%\begin{algorithm}n
%\caption{Distribution-based Adaptive Clustering Algorithm (DACA)}\label{algorithm}
%\KwData{initial the connected graph ${{{\cal G}}_0} = \left( {{{\cal K}},{{{\cal E}}_0}} \right)$ \;
%for $k$ in ${\cal K}$ : Compute ${D_{EMD}}$ by equation (\ref{Eq:2}) \;
%${{\cal M}} = \emptyset $, ${{\cal C}} = \emptyset $, ${{\bf{N}}^s} = \emptyset $.
%}
%\KwResult{${\cal M}$, ${\cal C}$, ${{\bf{N}}^s}$}
%\While{${{\cal M}} \cup {{\cal C}} \ne {{\cal K}}$}
%{Choose $c  \in {{\cal K}}\backslash \left( {{{\cal M}} \cup {{\cal C}}} \right)$ with $\max {\rm{ }}{D_{EMD}}$\;
%Choose $m \in \{ m|e\left( {m,c} \right) = 1\}$ with $\min {\rm{ }}{D_{EMD}}$\;
%Calculate ${v_{m,c}}$ by equation (\ref{Eq:9})\;
%${{\cal M}} \leftarrow {{\cal M}} \cup \left\{ m \right\}$\;
%${{{\cal C}}_m} \leftarrow {{{\cal C}}_m} \cup \left\{ c \right\}$\;
%${{\cal C}} \leftarrow \left\{ {...,{{{\cal C}}_m},...} \right\}$\;
%Update ${{\bf{N}}^s}$ by $N_m^S = \mathop {\min }\limits_{c \in {{\cal C}_m}} \left\{ {\frac{{{T_{th}}}}{{{v_{m,c}}}},{n_m}} \right\}$.}
%\end{algorithm}
\begin{algorithm}
\small{
\caption{Distribution-based Adaptive Clustering Algorithm (DACA)}\label{algorithm}
\KwIn{the social closeness graph ${{\cal G}} = \left( {{{\cal K}},{{\cal E}}} \right)$; \\
the constraints ${e_{{\rm{th}}}}$, ${v_{{\rm{th}}}}$, ${T_{{\rm{th}}}}$.
}
\KwOut{${\cal M}$, ${\cal C}$, ${{\bf{N}}^s}$}
\For{$k \in \mathcal{K}$}{
\For{$j \in \mathcal{K}\backslash \left\{ k \right\}$}{
Compute ${v_{k,j}}$ by Eq. (\ref{Eq:9})\;
Compute ${D_{{\rm{EMD}}}}\left( {k,j} \right)$ by Eq. (\ref{Eq:15});
}
Compute ${D_{{\rm{EMD}}}}\left( k \right)$ by Eq. (\ref{Eq:1}) and sort it\;
}
Build the connected graph ${\cal G}' = \left( {{\cal K},{\cal E}'} \right)$, where {\small ${\cal E}' = \left\{ {{{\tilde e}_{k,j}}= {D_{{\rm{EMD}}}}\left( {k,j} \right)|{e_{k,j}} \ge {e_{{\rm{th}}}},{v_{k,j}} \ge {v_{{\rm{th}}}}} \right\}$ }\;
Select $\cal M$ in descending order ${D_{{\rm{EMD}}}}\left( k \right)$, making all nodes can be connected by cluster heads $\cal M$\;
\While{${{\cal M}} \cup {{\cal C}} \ne {{\cal K}}$}
{Choose $m \in {\cal M},c \in {{\cal K}}\backslash \left( {{{\cal M}} \cup {{\cal C}}} \right)$ with $\max \, {\tilde e_{m,c}}$\;
${{{\cal C}}_m} \leftarrow {{{\cal C}}_m} \cup \left\{ c \right\}$\;
${{\cal C}} \leftarrow \left\{ {...,{{{\cal C}}_m},...} \right\}$\;
Compute ${{\bf{N}}^s}$ by $N_m^S = \mathop {\min }\limits_{c \in {{\cal C}_m}} \left\{ {\frac{{{T_{th}}}}{{{v_{m,c}}}},{n_m}} \right\}$.}
}
\end{algorithm}

%For the sake of simplicity, forming a graph ${{\cal G}_0} = \left( {{\cal K},{{\cal E}_0}} \right)$ as initial input, where ${{\cal E}_0} = \{ {\tilde e_{k,j}} \in \left\{ {0,1} \right\},\forall k,j \in {\cal K}\} $ denotes the set of edges or connections between users. ${\tilde e_{k,j}} = 1$ when ${e_{k,j}} \ge {e_{th}}$ and vice versa. As shown in Algorithm \ref{algorithm}, we use a greedy strategy to connect cluster members and cluster heads until all users are selected. Firstly, calculate the ${D_{{\rm{EMD}}}}$ for all users and select the cluster member $c$ with the maximum ${D_{{\rm{EMD}}}}$ among the remaining unassigned users. Then select cluster head m with the minimum ${D_{{\rm{EMD}}}}$ among users who are connected to the cluster head $c$. Finally, update the data volume of clustered data sharing. Repeat the selection process until all users are assigned. This algorithm can adaptively determine the number of clusters and has low complexity.

\section{Experiments}\label{SimuRe}
\subsection{Experiment Setup}
\emph{1)} Data Setting: 100 users participate in a deep learning task of classifying 10 classes over the MNIST dataset, where each image has 784 pixels and a label. Each pixel consists of 8 bits, and each label consists of 4 bits. Thus, $a = 6276$ bits per sample. For the non-IID data setting, 95 users receive data partition from only a single class, while the remaining users select samples without replacement from the training dataset with all labels.

\emph{2)} Model Structure: We adopt CNNs model since CNNs are very popular in imagine classification tasks. Specifically, we adopt a 2-layer CNN with two $5 \times 5$ convolution layers and two fully connection layers. During local training, we set batch-size as 10, local epoch as 1, learning-rate as 0.01 in SGD optimizer, the maximum number of communication rounds as 1500.

\emph{3)} Wireless Communication: For the social closeness setting, we use a random symmetric matrix ${\left\{ {{e_{i,j}}} \right\}_{K \times K}}$ as connected relationship. Considering bad, fair and good communication conditions, we set the SNR = -30 dB, 10 dB and 30 dB. We set the D2D bandwidth of each user as $B_{m}^s=100$ KHz and the wireless channels are modeled as the classic Raleigh fading channels. A round of FL takes 0.7s to complete according to measurement and computation (including uploads and local updates).

\textbf{Baselines.} To demonstrate the performance improvement, the clustered data sharing method is compared with two strategies considering maximized social closeness or data volume.
%\addtolength{\topmargin}{0.04in}

\begin{itemize}
\item FedAvg \cite{3}: Traditional FL with a serve and distributed users to train a global model without data sharing.
\item FedAvg with Central Sharing \cite{9}: A portion of the global dataset is sent directly by the server to all users.
\item Social Closeness Clustering (SCC): Clustering the users by social closeness, and the most trusted users are selected as cluster heads.
\item Communication Efficient Clustering (CEC): Clustering the users by transmission rate, and the users with high multicast rates are selected as cluster heads.
\end{itemize}
\subsection{Experiment Result}
\begin{figure}[]
\begin{center}
\subfloat[\footnotesize{Loss on non-IID data}]{\includegraphics[width=6cm,height=3.3cm]{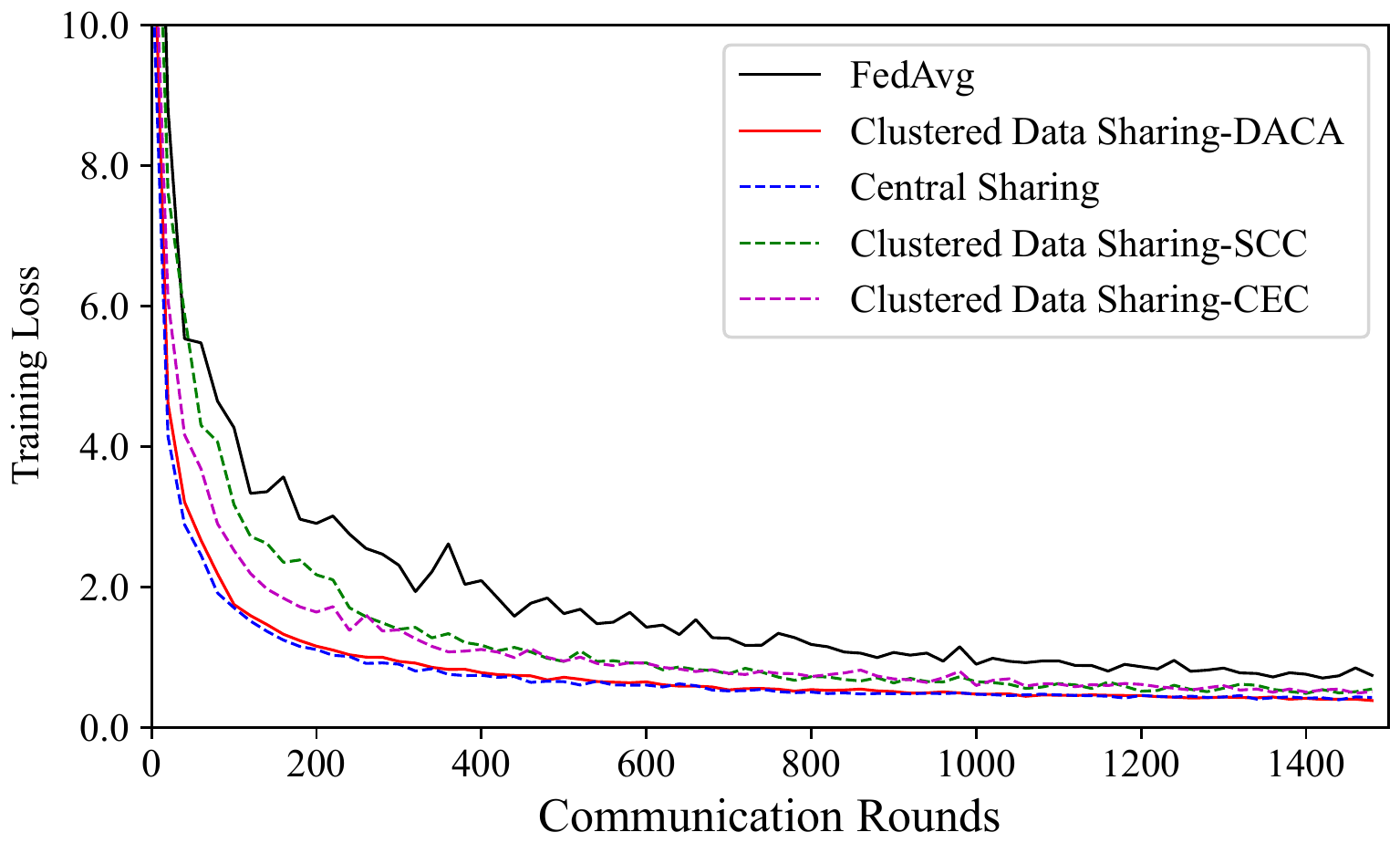}%
\label{fig3-1}}
\vspace{-4mm}
\quad
\subfloat[\footnotesize{Accuracy on non-IID data}]{\includegraphics[width=6cm,height=3.3cm]{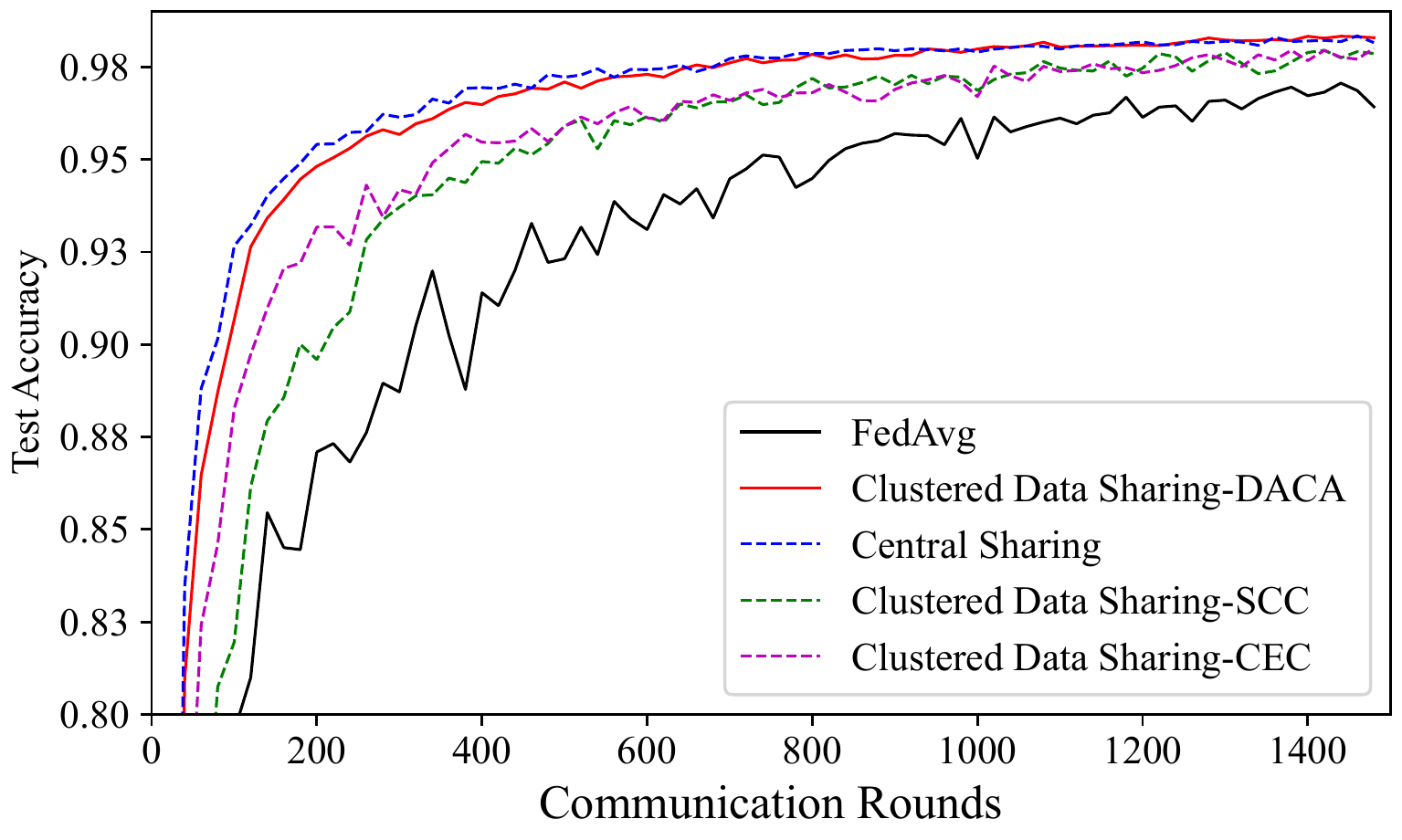}%
\label{fig3-2}}
\caption{The performance compared among baselines and ours, ${T_{th}}$ = 1s and ${e_{th}} = 0.5$.}
\vspace{-1.5mm}
\label{fig_3}
\vspace{-7mm}
\end{center}
\end{figure}

\emph{Performance for FL:} In Fig. \ref{fig3-1} and Fig. \ref{fig3-2}, the experiment results of training loss and test accuracy are provided. From Fig. \ref{fig3-1}, we can see that the proposed clustered data sharing methods can reduce the number of communication rounds from about 1000 to 400 compared to FedAvg, which is close to central sharing but better than other baselines. From Fig. \ref{fig3-2}, we can see that the proposed FL framework can improve the test accuracy from 96.42$\%$ to 98.72$\%$ compared to FedAvg. These gain terms from the fact that the proposed FL framework reduces the degree of non-IID data, and modifies the deviation of model update by SGD.

\begin{figure}[ht]
\begin{center}
\includegraphics[width=6cm,height=3.8cm]{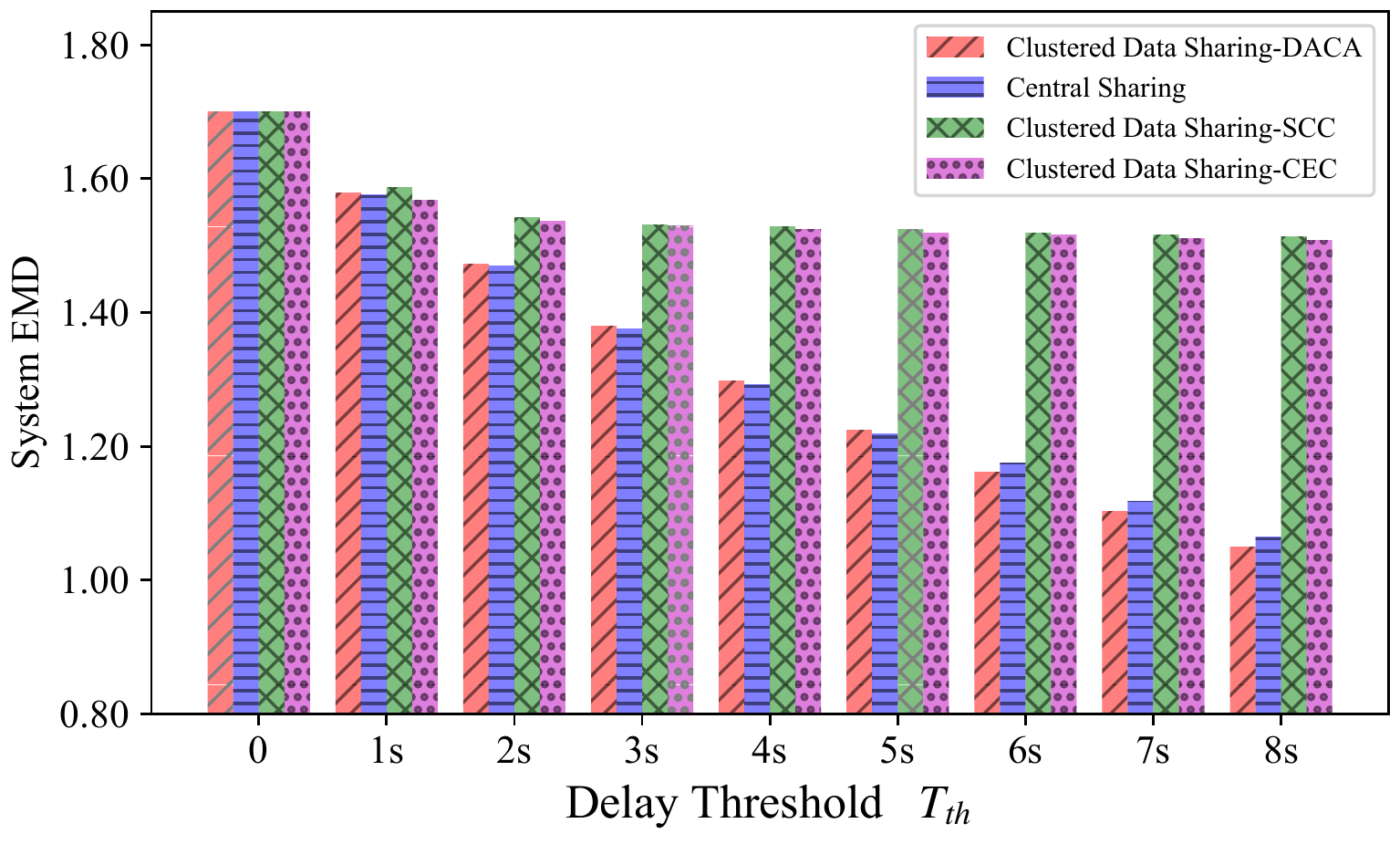}
\caption{The EMD under different ${T_{th}}$ and ${e_{th}} = 0.5$.}
\vspace{-6mm}
\label{fig_4}
\end{center}
\end{figure}
\emph{Effect of data volume:}  In Fig. \ref{fig_4}, we show how the system EMD changes as the delay threshold of data sharing varies. As shown in the figure, the value of system keeps decreases as the delay threshold increases. This is because data sharing makes the data distribution across devices more similar. Furthermore, as the delay threshold continues to increase, the system EMD values of baselines SCC and CEC change little, but our method DACA remains effective. It is reasonable because these clustering strategies do not consider the data distribution of devices and would share unfavorable data to contaminate the data distribution of some low EMD devices.

\begin{figure}[ht]
\begin{center}
\includegraphics[width=6cm,height=3.8cm]{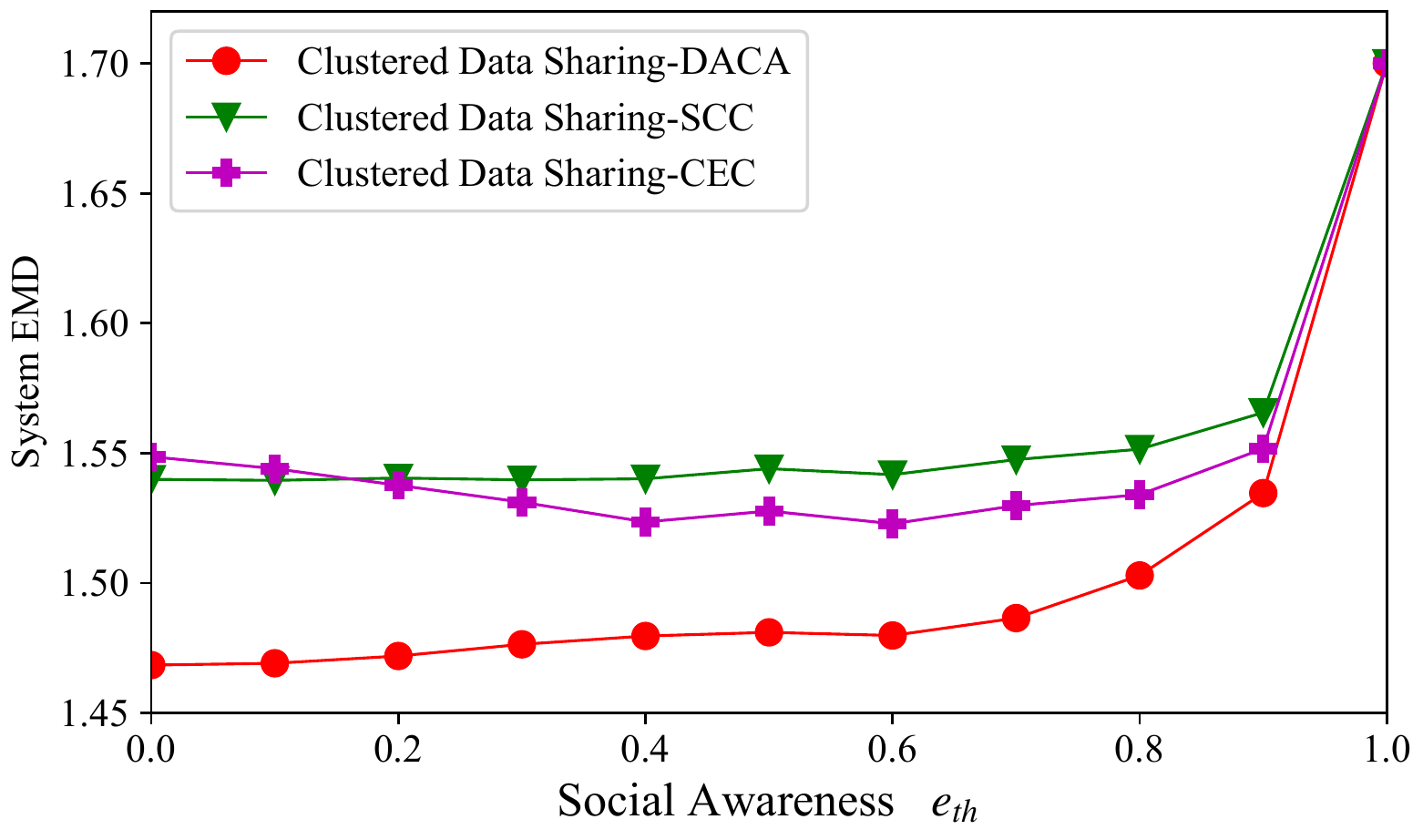}
\caption{The influence of social awareness on system EMD and ${T_{th}}$ = 2s.}
\vspace{-6mm}
\label{fig_5}
\end{center}
\end{figure}

\emph{Effect of social awareness:} Fig. \ref{fig_5} shows how the system EMD changes as the social awareness threshold varies. In this figure, as the social awareness threshold is too large, the system EMD of all methods increases rapidly. This is due to the fact that cluster heads can only share data with a few trusted cluster members and low EMD devices play a restrictive role. Our method outperforms other clustering strategies with the same social awareness threshold. This is because our method  considers the data distribution of cluster heads in addition to social awareness and transmission rate.

\begin{figure}[ht]
\begin{center}
\includegraphics[width=6cm,height=3.8cm]{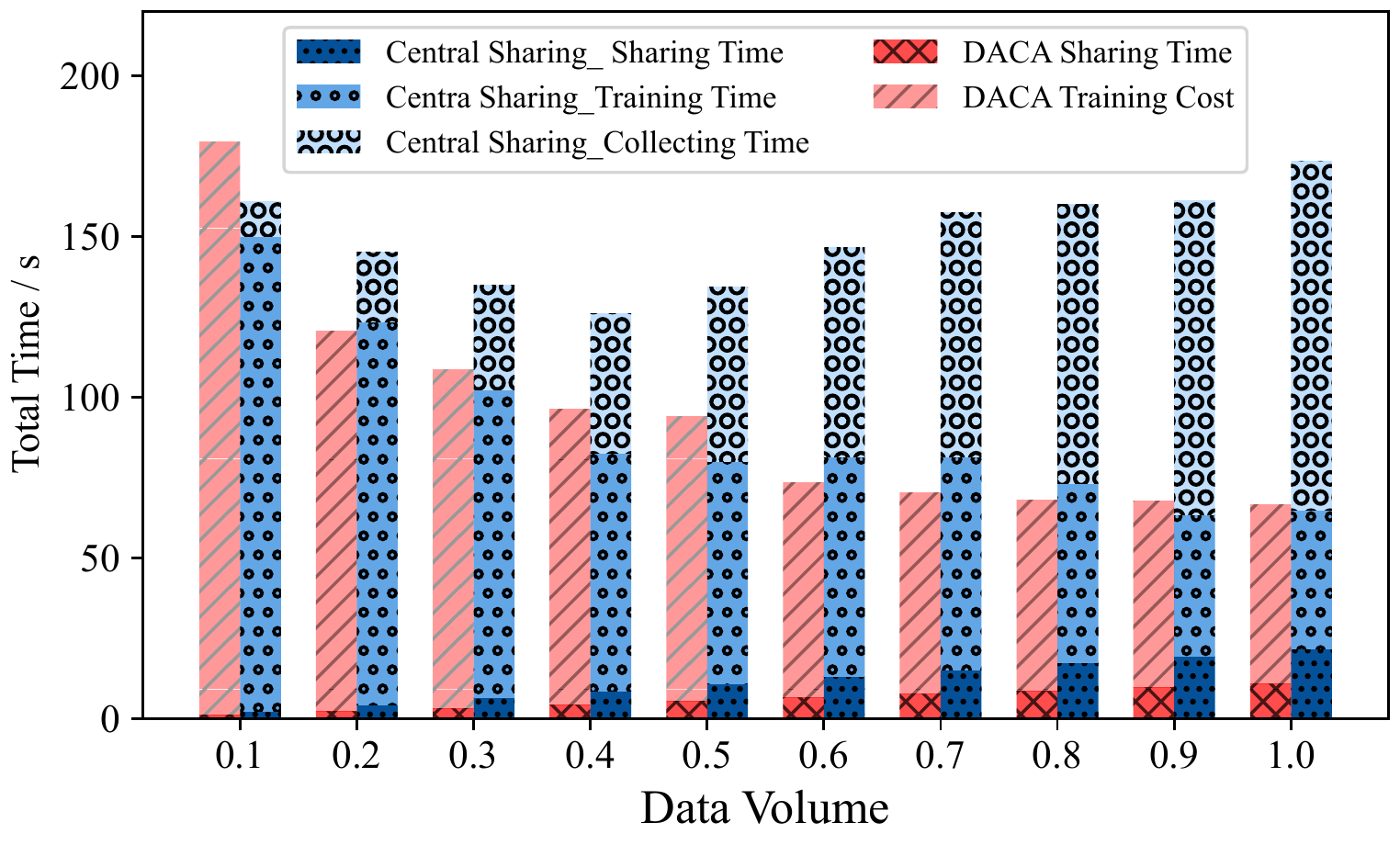}
\caption{The communication costs of FL with data sharing.}
\vspace{-6mm}
\label{fig_6}
\end{center}
\end{figure}

\emph{The Communication cost:} In Fig. \ref{fig_6}, we show how the total time varies with the data volume of data sharing, where data volume =1.0 is sharing 600 samples. For clustered data sharing FL, the total time of entire process decreases as the data volume increases. This is because our method can reduce the number of communication rounds in the FL training process. For central data sharing, the total time consists of data collecting, data sharing, and FL training. The total time of entire process first decreases and then increases as the data volume increases. This result may be explained by the fact that the time of FL training decreases but the time of collecting and sharing data increases. In summary, clustered data sharing FL can achieve almost identical benefits without the cost of data collection compared to central sharing.

\vspace{-1mm}
\section{Conclusion} \label{Clusion}
%\addtolength{\topmargin}{0.111in}
\vspace{-1mm}
In this work, we propose a clustered data sharing framework to solve the non-IID challenge in FL and design an optimization problem in wireless networks to minimize the system EMD. Through problem analysis, we derive three conditions to assist in clustering based on the privacy-preserving constrained graph. To maximize the gain from data sharing, a clustering algorithm is proposed to select the cluster heads and the credible associates based on the data distribution and the constraints. The experimental results show that our method performs well not only in terms of accuracy with limited transmission delay but also under privacy constraints.

\section*{Acknowledgment}

This work was supported in part by the National Key R \& D Program of China (No. 2021YFB3300100), the National Natural Science Foundation of China (No. 62171062).

\appendix
\emph{To prove the contrapositive of Theorem \ref{thm}:} \label{proof_1} Assume that ${\cal M}'$ and ${\cal C}'$ (${\cal M}' \ne {{\cal M}^*}$, ${\cal C}' \ne {{\cal C}^*}$) not satisfy condition 2, i.e., there exists a cluster member ${c_0} \in {{\cal C}'_{{m_0}}}$ whose ${D_{{\rm{EMD}}}}\left( {{c_0}} \right)$ is lower than its cluster heads ${m_0} \in {\cal M}'$. Then we can set ${c_0}$ as the cluster head for a new extra cluster, making the ${\tilde D_{{\rm{EMD}}}}$ lower. So ${\cal M}'$ and ${\cal C}'$ are not the optimal solutions of the problem $\mathcal{P}$. For ${\cal M}'$ and ${\cal C}'$ satisfy condition 2 but not satisfy condition 3, the data distributions of the cluster heads are all relatively close to the global distribution. Thus, the greater the difference in distribution distance between nodes and cluster heads, the greater the shared gain that can be obtained, making ${\tilde D_{{\rm{EMD}}}}\left( {{\cal M}',{\cal C}'} \right) > {\tilde D_{{\rm{EMD}}}}\left( {{{\cal M}^*},{{\cal C}^*}} \right)$. Hence, ${\cal M}'$ and ${\cal C}'$ are not the optimal solutions to the problem $\mathcal{P}$.

\bibliographystyle{plain}
\bibliography{IEEEexample}%参考文献文件名，不需要后缀

%\section*{References}
%
%\begin{thebibliography}{00}
%\bibitem{b1} G. Eason, B. Noble, and I. N. Sneddon, ``On certain integrals of Lipschitz-Hankel type involving products of Bessel functions,'' Phil. Trans. Roy. Soc. London, vol. A247, pp. 529--551, April 1955.
%\bibitem{b2} J. Clerk Maxwell, A Treatise on Electricity and Magnetism, 3rd ed., vol. 2. Oxford: Clarendon, 1892, pp.68--73.
%\bibitem{b3} I. S. Jacobs and C. P. Bean, ``Fine particles, thin films and exchange anisotropy,'' in Magnetism, vol. III, G. T. Rado and H. Suhl, Eds. New York: Academic, 1963, pp. 271--350.
%\bibitem{b4} K. Elissa, ``Title of paper if known,'' unpublished.
%\bibitem{b5} R. Nicole, ``Title of paper with only first word capitalized,'' J. Name Stand. Abbrev., in press.
%\bibitem{b6} Y. Yorozu, M. Hirano, K. Oka, and Y. Tagawa, ``Electron spectroscopy studies on magneto-optical media and plastic substrate interface,'' IEEE Transl. J. Magn. Japan, vol. 2, pp. 740--741, August 1987 [Digests 9th Annual Conf. Magnetics Japan, p. 301, 1982].
%\bibitem{b7} M. Young, The Technical Writer's Handbook. Mill Valley, CA: University Science, 1989.
%\end{thebibliography}
%\vspace{12pt}
%\color{red}
%IEEE conference templates contain guidance text for composing and formatting conference papers. Please ensure that all template text is removed from your conference paper prior to submission to the conference. Failure to remove the template text from your paper may result in your paper not being published.

\end{document}